# A Batch Normalization Classifier for Domain Adaptation

Matthew R. Behrend
Blue Well 8, LLC
behrend04@gmail.com

Sean M. Robinson
Pioneer Square Labs
sean@psl.com

*Abstract*— **Adapting a model to perform well on unforeseen data outside its training set is a common problem that continues to motivate new approaches. We demonstrate that application of batch normalization in the output layer, prior to softmax activation, results in improved generalization across visual data domains in a refined ResNet model. The approach adds negligible computational complexity yet outperforms many domain adaptation methods that explicitly learn to align data domains. We benchmark this technique on the Office-Home dataset and show that batch normalization is competitive with other leading methods. We show that this method is not sensitive to presence of source data during adaptation and further present the impact on trained tensor distributions tends toward sparsity. Code is available at https://github.com/matthewbehrend/BNC**

## I. INTRODUCTION

Neural networks can perform well on computer vision tasks for a specific distribution of data. Accuracy tends to diminish when presented with slightly unfamiliar data, such as images captured in different lighting [1], or with subjects in a different pose [2] than was typical of training samples. During training, certain data may have been unavailable or unanticipated that the model may encounter in deployment. Unsupervised domain adaptation (UDA) includes a variety of methods that attempt to align features between a source domain and a target domain to improve inference accuracy in an unlabeled target domain [1-3]. In addition to natural domain shifts, models are increasingly being trained on synthetic data, such as video game scenes [4] or other simplifications of the real world and using domain adaptation to perform in the real world [5, 6].

We address the problem of adapting a network with source-domain knowledge to perform its task in an unlabeled target domain without access to source-domain data (source-free UDA or SFUDA). Applications of UDA can include image classification [6], facial recognition [7], traffic scenes [1], sentiment analysis [8], and speech emotion recognition [9]. Some approaches to domain adaptation defer the learning problem to additional networks, such as a domain discriminator in adversarial domain adaptation [6-8]. Others duplicate the classifier network to be refined to the target domain [9, 10]. These may be source-free or co-trained with source data during adaptation. Even source-free methods usually require the adaptation process to have access to source-domain information such as feature prototypes or statistical distributions in the source domain.

Here we report on a new method for SFUDA based on batch normalization [11] following the last fully connected layer of a classifier. Batch normalization is a method of regularizing the mean and variance of data in the latent layers of a neural network to achieve higher learning rates. How it improves training is still only partially understood but recent work has shown a decoupling effect on weights [12] and smoothing of the optimization surface [13]. Our batch normalization classifier (BNC) is unique in that it works on the existing network in-place with little added complexity.

This paper contributes:

A batch normalization classifier for source-free unsupervised domain adaptation.

Demonstration of performance on Office-Home competitive with more complex methods.

Analysis of the effects of batch normalization on the distributions of features and trained weights.

## II. METHODS

### A. Architecture

Our BNC method adds a batch-normalization (BN) layer between the last fully-connected (FC) layer and the softmax (SM) activation of the classifier network. We used ResNet-50 [14] pretrained on ImageNet as the backbone for our model (Fig. 1). The ResNet-50 classifier layer was removed and its feature layer (2048 channels) was fed into a network of two fully connected layers, composed as listed in Table 1. The number of classes, k, was 65 for the network trained on Office-Home data.

The loss function for supervised training in the source-domain data was cross entropy (Eq. 1) to the one-hot label vector. Adaptation in the target domain was performed without labels, using entropy (Eq. 2) as the loss function. Entropy minimization is a common technique for unsupervised separation of data [15, 16].

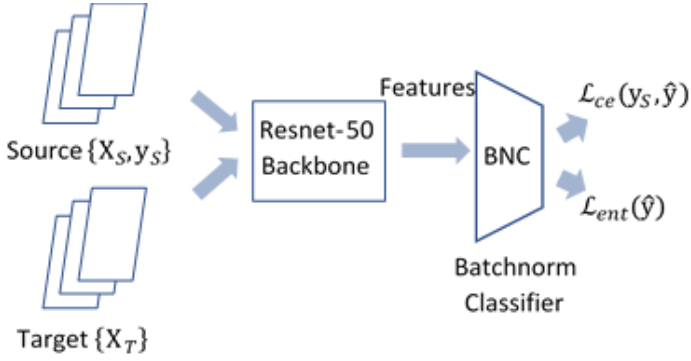

Fig. 1. The classifier was trained with cross-entropy loss on labeled source-domain images and with entropy loss on unlabeled target-domain images. Features were pre-extracted using the ResNet-50 backbone pretrained on ImageNet.

TABLE I. BATCHNORM CLASSIFIER REFINEMENT OF RESNET-50

| Layer Type | Layer Name | Output Size |
|---|---|---|
| Resnet-50 Image Features | Backbone | N x 2048 x 1 x 1 |
| Flatten | FL | N x 2048 |
| Fully Connected, 512 | FC_1 | N x 512 |
| Leaky ReLU, leak=0.2 | LR_1 | N x 512 |
| BatchNorm | BN_1 | N x 512 |
| Dropout, p=0.5 | DO_1 | N x 512 |
| Fully Connected, 65 | FC_2 | N x k |
| BatchNorm | BN_2 | N x k |
| SoftMax | SM | N x k |

$$\mathcal{L}_{ce}(y_S, \hat{y}) = \frac{1}{N} \sum_{i=1}^{N} \sum_{c=1}^{k} -y_{S,i,c} \log(\hat{y}_{i,c}) \quad (1)$$

$$\mathcal{L}_{ent}(\hat{y}) = \frac{1}{N} \sum_{i=1}^{N} \sum_{c=1}^{k} -\hat{y}_{i,c} \log(\hat{y}_{i,c}) \quad (2)$$

## III. EXPERIMENTS

### A. Dataset

We performed source-free unsupervised domain adaptation of our BNC model using the Office-Home dataset [17], a challenging dataset for domain adaptation containing images of 65 common objects in four domains: Art (Ar), Clipart (Cl), Product (Pr) and Real-World (Rw).

### B. Training

The pretrained Resnet-50 model was loaded from the MXNet Gluon model zoo. For each of 12 possible domain shifts among the four domains of Office-Home, the model was trained with supervision in the source domain and subsequently adapted in the target domain without labels. Accuracy was measured in the target domain. No source-domain data were made available during adaptation in the target domain.

Data were loaded in minibatches of 256 images with shuffling. The model was trained for five epochs in the source domain and then adapted for five epochs in the target domain. Each domain shift task was run three times and the average accuracy reported)

### C. Results

BNC performed better than many methods of greater complexity (Table 2), including methods making use of a domain discriminator. Despite relying only upon knowledge inherent to the network, the BNC was competitive with many leading methods, achieving an overall accuracy of 68.2% on the Office-Home benchmark.

BNC was quite insensitive to source domain data. Most UDA methods rely on co-training with source and target domain data, but BNC performance was not affected the presence or absence of source domain data (Table 3). Individual domain shifts within Office-Home (Table 3) were affected on the same order as training stochasticity

TABLE II. CLASSIFICATION ACCURACY ON OFFICE-HOME DATASET FOR UNSUPERVISED DOMAIN ADAPTATION

| Method | Source Free | Ar→Cl | Ar→Pr | Ar→Rw | Cl→Ar | Cl→Pr | Cl→Rw | Pr→Ar | Pr→Cl | Pr→Rw | Rw→Ar | Rw→Cl | Rw→Pr | Avg |
|---|---|---|---|---|---|---|---|---|---|---|---|---|---|---|
| HAFN [18] | n | 50.2 | 70.1 | 76.6 | 61.1 | 68 | 70.7 | 59.5 | 48.4 | 77.3 | 69.4 | 53 | 80.2 | 65.4 |
| CDAN+E [19] | n | 50.7 | 70.6 | 76 | 57.6 | 70 | 70 | 57.4 | 50.9 | 77.3 | 70.9 | 56.7 | 81.6 | 65.8 |
| SAFN [18] | n | 52 | 71.7 | 76.3 | 64.2 | 69.9 | 71.9 | 63.7 | 51.4 | 77.1 | 70.9 | 57.1 | 81.5 | 67.3 |
| BA3US [20] | n | 51.2 | 73.8 | 78.1 | 63.3 | 73.4 | 73.6 | 63.3 | 54.5 | 80.4 | 72.6 | 56.7 | 83.7 | 68.7 |
| CADA [21] | n | 56.9 | 76.4 | 80.7 | 61.3 | 75.2 | 75.2 | 63.2 | 54.5 | 80.7 | **73.9** | **61.5** | 84.1 | 70.2 |
| PrDA [22] | y | 48.4 | 73.4 | 76.9 | 64.3 | 69.8 | 71.7 | 62.7 | 45.3 | 76.6 | 69.8 | 50.5 | 79 | 65.7 |
| SHOT [23] | y | 57.1 | **78.1** | 81.5 | **68** | **78.2** | **78.1** | **67.4** | 54.9 | **82.2** | 73.3 | 58.8 | **84.3** | 71.8 |
| BAIT [24] | y | **57.4** | 77.5 | **82.4** | **68** | 77.2 | 75.1 | 67.1 | **55.5** | 81.9 | **73.9** | 59.5 | 84.2 | 71.6 |
| **BNC (ours)** | y | 51.3 | 74.7 | 78.1 | 65.2 | 77.5 | 77.4 | 64.1 | 52.3 | 78.8 | 66.9 | 53.6 | 78.6 | 68.2 |

TABLE III. BNC PERFORMANCE SENSITIVITY TO SOURCE DOMAIN DATA

| Method | Ar→Cl | Ar→Pr | Ar→Rw | Cl→Ar | Cl→Pr | Cl→Rw | Pr→Ar | Pr→Cl | Pr→Rw | Rw→Ar | Rw→Cl | Rw→Pr | Avg |
|---|---|---|---|---|---|---|---|---|---|---|---|---|---|
| BNC (source co-trained) | 50.1 | 73.7 | 76.8 | 65.5 | 75.3 | 75.9 | 64.2 | 51.3 | 78.8 | 67.4 | 53.2 | 78.5 | 67.6 |
| BNC | 51.3 | 74.7 | 78.1 | 65.2 | 77.5 | 77.4 | 64.1 | 52.3 | 78.8 | 66.9 | 53.6 | 78.6 | 68.2 |

## IV. ANALYSIS

### A. Feature Visualization

The approach developed in this work of utilizing the BN layer immediately prior to softmax is not commonly applied to classification problems. However, the addition of the BN layer is expected to relax some of the requirements otherwise imposed upon the FC network structure to achieve high accuracy. If one channel has a consistently lower output score, the BN layer may allow the highest-scored result in that channel to result in a classification of that channel, ultimately allowing a correct result without the need for similar inter-channel output distributions. In informal terms, the combination of the BN and SM layers allow the network to produce accurate classifications, even if "an image of a chair" does not produce an absolutely higher score in the "chair "channel than every other incorrect channel - if it is relatively above the norm in the chair channel more than it scores above the norm in other classification channels. This in turn allows the FC layers only to encode meaningful relationships.

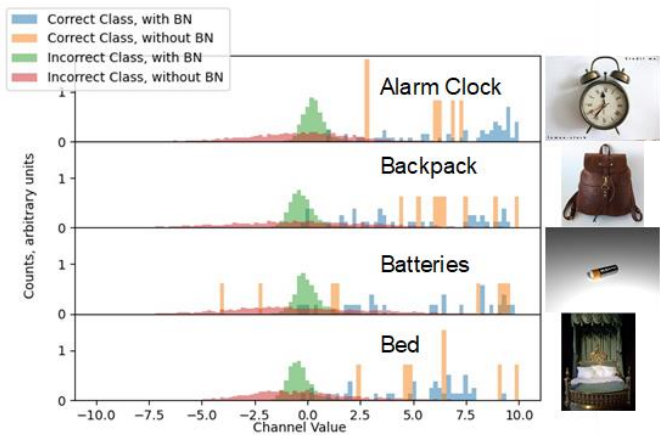

Fig. 2. Example channel-output histograms from the output of the FC network, first 4 channels.

With regard to training, the non-BN method appeared to train the tensors in a way that produced similar inter-channel output distributions. Outputs were broader (but overlapping) from the FC network layers without the presence of the BN layer (Fig. 2)

One potential explanation for this behavior could be that due to the value of similarity in output distribution between channels (for achieving classification accuracy), model training steps without the BN layer tend to add weight to the interaction of meaningfully unrelated vector elements to achieve this similarity. If so, this would effectively broaden the channel value distributions pre-softmax, and lead to larger average values for FC-layer tensor weights as well as less sparsity in those tensors. However, with the BN layer obviating this need for equivalence of distribution, the FC layers only experience pressure to yield higher intra-channel scores for correct classifications than incorrect ones), so tensor elements that connect a previous layer's output to a meaningfully unrelated channel will trend toward 0 under training. Likewise, we see deviation in inter-channel distributions when the BN layer is activated, but a correspondingly better relative separation between values for correct-channel and other-channel image classes.

### B. Ablation Study

We investigated the expected observations above, focusing on the outputs of the last fully-connected layer, and the inputs into softmax layer, with and without the presence of the BN2 layer. Fig. 3 shows the normalized histograms of the FC2 layer output, and the input to the SM layer, with and without the presence of a BN layer between them. Note that in the absence layer BN2, these two plots are the same. However, without the BN layer, a broadening of the scores created more overlap between the scores of correct classifications and incorrect classifications. With BN2 layer in place, incorrect classification scores were narrowed (green versus red counts in Fig 3), suggesting a shift in meaningful nonzero terms in the FC weights, allowing better separation between correct and incorrect classes.

To demonstrate the narrower and better-separated distributions achieved with the BN method, we employed a reduced-dimensionality projection using t-SNE [25] on the inputs to the softmax layer (Fig. 5). The method with the BN layer before the classification layer shows an improved separation between image classes.

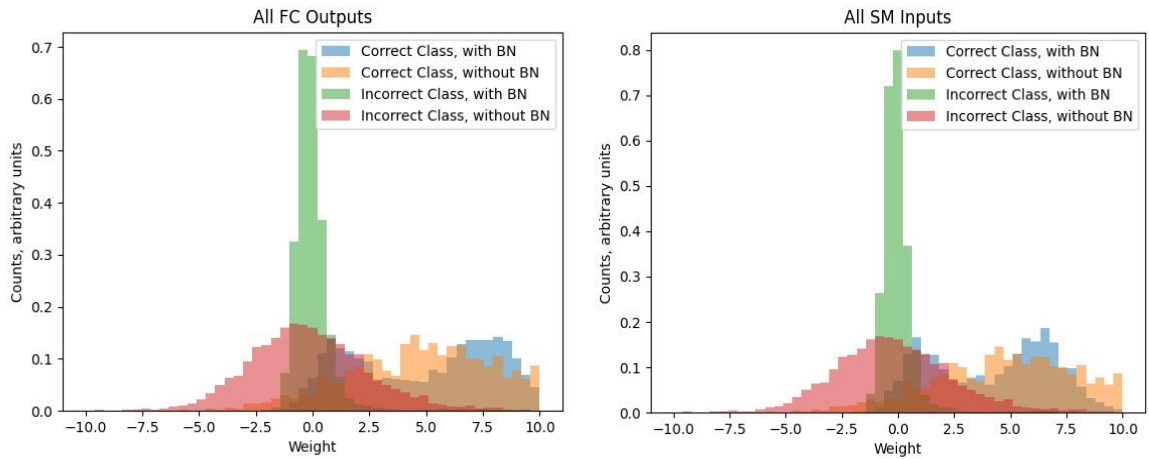

Fig. 3. Histograms of fully-connected (FC) outputs and softmax (SM) input (all channels)

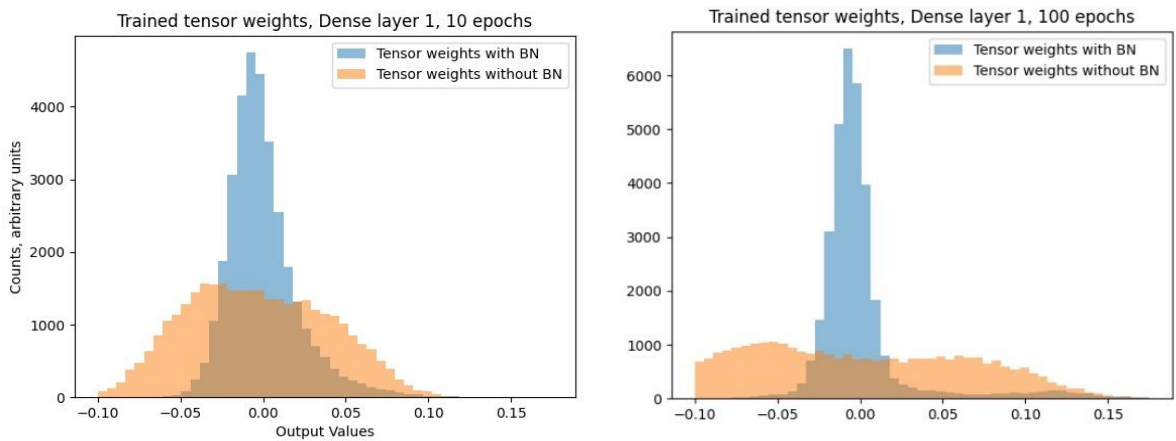

Fig. 4. Histogram of trained FC-layer weights

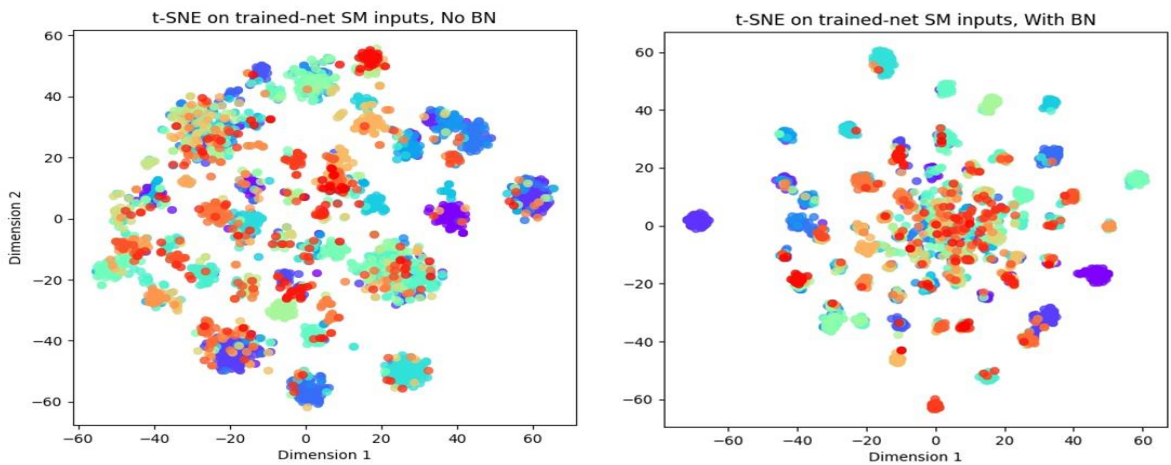

Fig. 5. 2D t-SNE projections of the inputs to the softmax layer

## V. DISCUSSION

We have demonstrated a simple approach to SFUDA based on batch normalization that is competitive with other state-of-the-art methods. It performs better than many methods of greater complexity including those with access to source-data. BNC is unique in its use of network source-domain knowledge without referencing a fixed copy of the source-trained model for pseudo-labels or class prototypes.

Other leading methods for SFUDA use some combination of entropy minimization [22, 23], pseudo-labeling [23], and class prototypes [22, 24]. PrDA [22] applies greater weight to samples that have lower entropy in the source-trained model or have greater separation from other class prototypes. PrDA uses the fixed source-trained model to generate pseudo-labels as well as class prototypes. SHOT [23] uses class prototypes to select nearest pseudo-labels from the source-trained model. BAIT splits the target data according to an entropy threshold and applies labels according to class prototypes from the source-trained model for confident samples and prototypes from the target-adapted model for uncertain samples. Each is a variation on the same theme of retaining a copy of the source-trained model as a hypothesis. Additional relevant approaches have been demonstrated on earlier datasets such as Office-31 and digits datasets [26-29] that we did not compare against.

We found that the addition of BN in the classifier had the effect of reducing the magnitude of features for incorrect classifications. We observed a better relative separation between channel values for correct and incorrect image classes, made possible by rescaling the output of the FC layer prior to softmax. Comparing feature separation to [20] we also observed a t-SNE pattern having a central region of poorly differentiated classifications surrounded by well differentiated classes with occasional confident misclassifications. Since we do not use a frozen source-trained model for label hypotheses, our approach may rely upon different mechanisms that are more closely related to smoothing the loss surface across domains, or otherwise encouraging robustness in the model that generalizes over domain shifts. We anticipate further advances in domain adaptation with hypothesis-based methods, potentially in combination with normalization methods.

Our model adds to recent experience that access to source-domain data may not be necessary for adaptation to a target domain. Better-performing UDA methods tend to make use of source-domain knowledge within the network [23] rather than re-exposing the network to source-domain data.